\documentclass[letterpaper]{article} 
\usepackage{aaai24}  
\usepackage{times}  
\usepackage{helvet}  
\usepackage{courier}  
\usepackage[hyphens]{url}  
\usepackage{graphicx} 
\def\UrlFont{\rm}  
\usepackage{natbib}  
\usepackage{caption} 
\usepackage{algorithm}
\usepackage{algorithmic}
\usepackage{multirow}
\usepackage{lipsum} 
\usepackage{newfloat}
\usepackage{listings}
\DeclareCaptionStyle{ruled}{labelfont=normalfont,labelsep=colon,strut=off} 
\floatstyle{ruled}
\newfloat{listing}{tb}{lst}{}
\floatname{listing}{Listing}
\usepackage{newfloat}
\usepackage{xcolor}%
\usepackage{tikz}
\def\checkmark{\tikz\fill[scale=0.4](0,.35) -- (.25,0) -- (1,.7) -- (.25,.15) -- cycle;} 

\title{Fusing Domain-Specific Content from Large Language Models into Knowledge Graphs  for Enhanced Zero Shot Object State Classification}

\author {
    Filippos Gouidis\textsuperscript{\rm 1,\rm 2,\rm 3},
    Katerina Papantoniou\textsuperscript{\rm 3},
    Konstantinos Papoutsakis\textsuperscript{\rm 2},
    Theodore Patkos\textsuperscript{\rm 3}, \\
    Antonis Argyros\textsuperscript{\rm 2,\rm  3} and
    Dimitris Plexousakis\textsuperscript{\rm 2,\rm  3}
}
\affiliations {
  \textsuperscript{\rm 1}Computer Science Department, University of Crete, Heraklion, Greece
  
    \textsuperscript{\rm 2}Department of Management, Science and Technology, Hellenic Mediterranean University, Agios Nikolaos, Greece \\
    \textsuperscript{\rm 3}Institute of Computer Science, Foundation for Research and Technology Hellas, Heraklion, Greece \\
         gouidis@\{ics.forth.gr, csd.uoc.gr, hmu.gr\}, kpapoutsakis@hmu.gr, \{papanton, patkos, argyros, dp\}@ics.forth.gr
}

\begin{document}


\maketitle

\begin{abstract}

Domain-specific knowledge can significantly contribute to addressing a wide variety of vision tasks. However, the generation of such knowledge entails considerable human labor and time costs. This study investigates the potential of Large Language Models (LLMs) in generating and providing domain-specific information through semantic embeddings. To achieve this, an LLM is integrated into a pipeline that utilizes Knowledge Graphs and pre-trained semantic vectors in the context of the Vision-based Zero-shot Object State Classification task. We thoroughly examine the behavior of the LLM through an extensive ablation study. Our findings reveal that the integration of LLM-based embeddings, in combination with general-purpose pre-trained embeddings, leads to substantial performance improvements. Drawing insights from this ablation study, we conduct a comparative analysis against competing models, thereby highlighting the state-of-the-art performance achieved by the proposed approach.

\end{abstract}

\section{Introduction}

The deep learning paradigm has revolutionized the standard approach concerning AI-based problem-solving in several scientific disciplines over the last few years, leading to spectacular breakthroughs. However,  the growing utilization and customization of these types of methods brought to light several limitations, the most important of which are (a)~their total dependency on significant quantities of annotated data whose collection and annotation is a costly and time-consuming procedure, and (b)~the vast computational cost entailed by these approaches. Therefore, a great deal of research has been dedicated in recent years to the development of methods to overcome these limitations.  

General-purpose semantic representations, typically in the form of pre-trained word vectors, can be considered as an answer to these challenges since they offer several attractive features. First, they are readily available, straightforward to use, and can be easily integrated in many different frameworks that address a wide variety of tasks. Equally important, in certain cases,  performance gains can be obtained by their utilization, especially when used in conjunction with Knowledge Graphs (KGs) which enable a structured and robust way to represent information concerning entities and relationships, and serve as powerful tools for organizing and connecting diverse knowledge in a networked fashion. Nevertheless, even though these general-purpose representations are often trained on vast corpora, they still lack coverage for specialized topics, while  it also remains difficult to handle polysemy. Such representations can provide an average representation based on the different meanings of a word, but this may not suffice in specific problems.

On the other hand,  
domain-specific knowledge can mitigate the aforementioned limitations. However, the generation of such knowledge requires typically human expertise and considerable amounts of time and effort. The advent of Large Language Models (LLMs) seems to offer an efficient way to overcome these hurdles since models of this type can provide expertise comparable to that of humans in a fast and semi-automated way. The utilization of  LLMs poses other challenges, the most stressing of which concern their effective prompting, their struggle with commonsense knowledge and reasoning, and their difficult discernible biases~\cite{Sartori2023cognitive}.

In light of these points, significant research in various fields is currently concentrated on performing prediction tasks on samples for which no prior training has been performed, an approach that is typically referred to as zero-shot learning. Computer Vision is one of the fields, where zero-shot methods are extensively investigated. 
One reason for this is the already mentioned cost for data collection and annotation.  Another important advantage of the zero-shot-based methods is that standard methods, when utilized in datasets with different distributions from the ones that have been trained, typically suffer from performance degradation. However, zero-shot methods are impervious to this kind of limitation. Nevertheless, the development of robust zero-shot methods presents a formidable challenge with its own set of unique hurdles, the most important of which are the \textit{retrieval} of  pertinent information that will be used by the models as a proxy to the ``missing'' visual information and the effective \textit{projection} of this information to the visual space, enabling its utilization for the associated vision task.

In this study, we consider the problem of vision-based Object State Classification (OSC), a task involving the recognition of object states in images. Efficient OSC is crucial for any agent interacting with objects, given the profound impact states can exert on overall object functionality. OSC presents several significant challenges that remain unresolved and is currently a focal point of increasing interest within the research community. Specifically, our attention is directed toward the challenging variant of Object-Agnostic zero-shot OSC, which aims to deduce the state of an object without relying on visual information related to the state classes or any general information about object classes, as described in~\cite{gouidis204}.
Building upon recent research addressing the specified problem~\cite{nayak2022learning, gouidis204}, we extend the state-of-the-art by integrating LLM frameworks into the broader pipeline designed for addressing Object-Agnostic zero-shot OSC, investigating appropriate schemes for an efficient coupling of the components. A key feature of our methodology is the utilization of an LLM that generates domain-specific knowledge relevant to the task.

To investigate this approach, we conduct a thorough ablation study to explore the optimal way in which the information derived from the LLM can be effectively leveraged. The experimental results indicate that the fusion of this domain-specific knowledge with pre-trained general-purpose knowledge has the potential to achieve state-of-the-art (SoA) performance in the Object-Agnostic zero-shot OSC task.

We believe that the insights gained from our approach extend beyond the specific problem under examination and can prove advantageous for other challenges associated with zero-shot learning. Additionally, it is essential to emphasize that our work represents an initial stride towards creating an effective framework that seamlessly integrates Large Language Models (LLMs) into a hybrid method, incorporating both low-level (data-driven) and high-level (symbolic) components.

The preliminary results presented in this work instill confidence in the potential of this approach. We view it as a promising step towards the development of an efficient framework that can serve as a foundation for addressing a broader spectrum of problems, particularly those within the realm of zero-shot learning.

In summary, this paper makes the following contributions:

\begin{itemize} 
\item We introduce a novel approach that integrates LLMs into the successful zero-shot framework for object state classification which combines Knowledge Graphs and pre-trained semantic vectors.
\item We present a semi-automatic method that harnesses the general-purpose content of an LLM to generate domain-specific knowledge.
\item Through an extensive ablation study, we explore various aspects of the optimal integration of our method.
\item Building upon the insights gained from the ablation study, we compare our method against competing models, demonstrating its state-of-the-art performance.
\end{itemize}
\section{Related Work}
\vspace*{0.1cm}\noindent\textbf{Leverage LLMs}: 
The integration of LLMs into tasks involving KGs embedding is an active area of research, as demonstrated by \cite{pan2023unifying}. \citep{wang-etal-2021-kepler} introduces KEPLER, a unified model for knowledge embedding and pre-trained language representation. This model encodes textual entity descriptions into embeddings using a pre-trained model, achieving joint training objectives for knowledge embedding (KE) and masked language modeling (MLM).
In another study, \citep{huang2022endowing} combines LLMs with vision and graph encoders to generate multimodal KG embeddings. Evaluation on multilingual named entity recognition and visual sense disambiguation tasks indicates improved performance compared to alternative approaches.
LLMs have also been applied to assist open-vocabulary models for zero-shot image classification, as seen in the work of \citet{Pratt2022WhatDA}. This approach involves constructing prompts specialized for the target dataset for the GPT-3 model, with the goal of generating text containing crucial discriminating characteristics of image categories.


\vspace*{0.1cm}\noindent\textbf{Fusion of word embeddings}:
Various approaches for the fusion of embeddings in Natural Language Processing have been explored. In \citep{Ghannay2016Word}, a comprehensive evaluation of different word embeddings (e.g., CBOW, Glove, Skip-gram, w2vf-deps) demonstrates performance improvements through their combination, emphasizing complementarity. The fusion of generic and domain embeddings, closely aligned with the objective of this work, is investigated in \citep{Sarma2018domain} and \citep{Seyler2020Domain}.
In \citep{Sarma2018domain}, an approach aligns word vectors using Canonical Correlation Analysis and non-linear Kernel Canonical Correlation Analysis \cite{Fukumizu2007KCCA}, followed by combining generic and domain embeddings through convex optimization. \citep{Seyler2020Domain} explores three fusion approaches: concatenation of embedding vectors, weighted fusion of text data, and interpolation of aligned embedding vectors, with the interpolation method yielding optimal results. Moreover, \citep{Rettig2019Fusion} suggests relying on domain-only embeddings for specific applications, proposing a two-step process. The first step involves ranking to capture the similarity between the downstream application corpus and various domain-specific embeddings. The second step employs fusion-based approaches, with Principal Components Analysis (PCA) over concatenated embeddings identified as the most efficient method.

\vspace*{0.1cm}\noindent\textbf{State Classification}:
Few works  address  state  classification specifically. The majority of the works related to the task adopts assumptions akin to those used in attribute classification tasks  \cite{gouidis2022}. 
A notable recent contribution is a multi-task, self-supervised learning method proposed by \cite{soucek2022multitask} which jointly learns to temporally localize object state changes and the corresponding state-modifying actions in videos.
A significant research focus in this domain is zero-shot learning, gaining attention for its practical relevance in real-world applications, especially in overcoming challenges associated with collecting and learning training data for a large number of object classes \cite{xian2018zero}. An influential approach in zero-shot learning involves using semantic embeddings to represent objects and their attributes in a low-dimensional space \cite{Wang2018b}.  The emergence of powerful generative models has opened up a promising avenue for zero-shot object classification \cite{xian2018feature}. In a more recent work, \cite{gouidis204} specifically tackles the challenging task of Object-Agnostic zero-shot State Classification, employing a strategy that leverages Knowledge Graphs and pre-trained semantic vectors.

\vspace*{0.1cm}\noindent\textbf{Vision-Language Models}: 
Vision-Language Models (VLMs) are sophisticated language models that harness contrastive learning to bridge the gap between images and text. They are essentially an extension of LLMs, trained on massive text data, tailored to tackle Computer Vision challenges. VLMs achieve this by jointly training image and text encoders on vast datasets of image-text pairs gleaned from the internet. This synergistic approach empowers the encoders to excel in downstream tasks like Image Captioning, Visual Question Answering, and Zero-Shot Classification. Notable examples of VLMs include CLIP~\cite{radford2021learning},  ALIGN~\cite{jia2021scaling} and BLIP~\cite{li2022blip}.


\section{Methodology}

Given a set of objects, $O$,     a set of states, $S$, and a set of images, $I$, with each image, $i \in I$ containing an object $o \in O$ that is in a state $s \in S$, the goal of OSC is to predict the state label $s \in S$, given an object $o$ appearing in an image $i \in I$ as input. In the zero-shot variation of OSC that we examine in this work,  no images are used for the training of the classifier model that is deployed for the prediction.

\subsection{Approach}

\begin{figure*}
\centering
\includegraphics[scale=0.18]{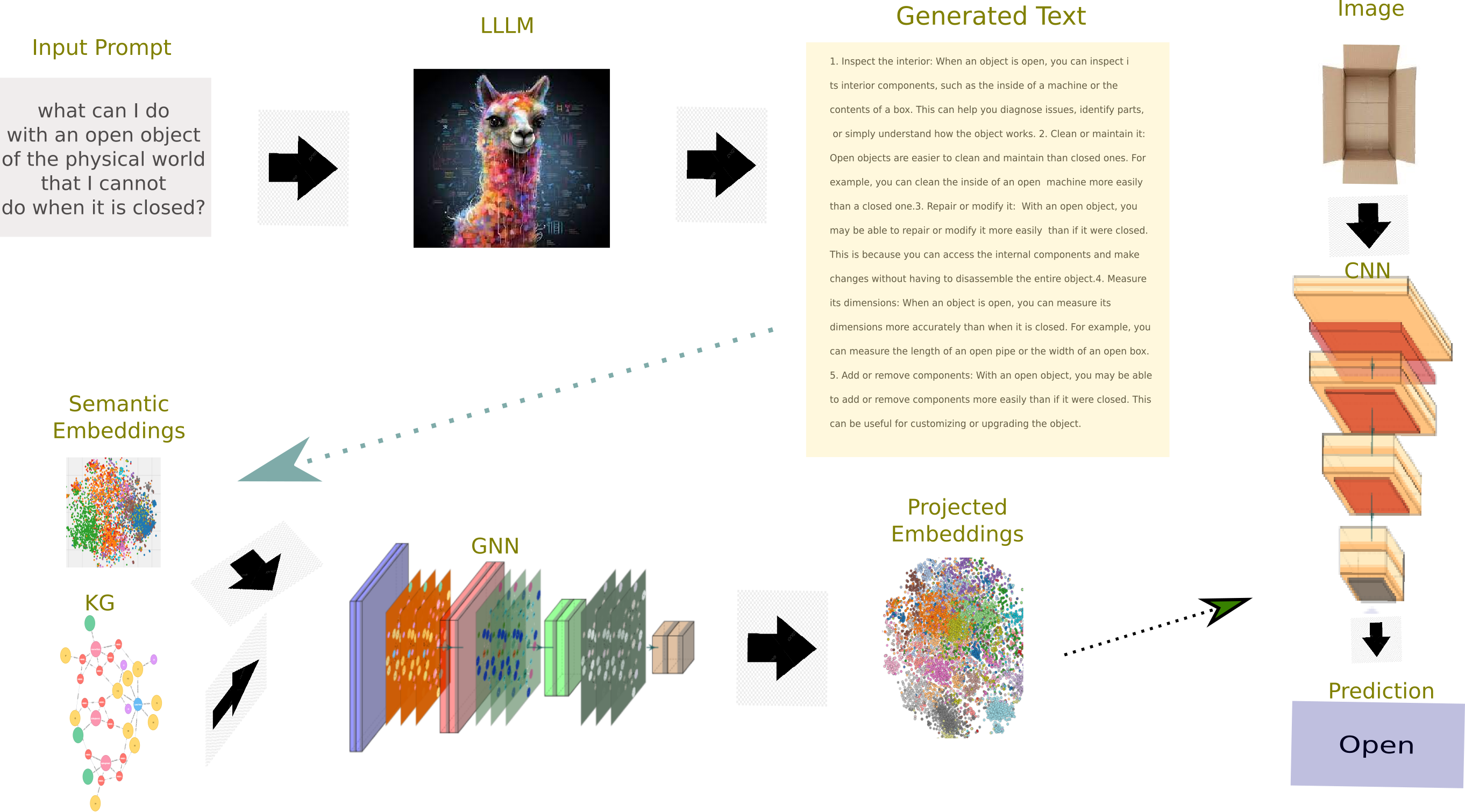}
  \caption{The schematic representation of our methodology. A Large Language Model (LLM) is given specific prompts associated with the target classes we aim to identify. The resulting corpus is processed, leading to the generation of semantic embeddings. Subsequently, these vectors are fed into a Graph Neural Network (GNN), previously trained to map embeddings from the semantic space to the visual space. The resulting visual embeddings are then integrated into the final layer of a pre-trained Convolutional Neural Network (CNN) classifier.}
  \label{fig:Overview}
\end{figure*}

Our approach consists of the following stages which are graphically illustrated in Figure~\ref{fig:Overview}: 
\begin{enumerate}
  \item Prompting of the LLM and generation of a text corpus.
  \item Construction of a KG relevant to the classes we are interested in recognizing. 
\item Production of semantic and visual embeddings.
  \item Training of a GNN that learns to project semantic embeddings into visual space.
  
  \item Projection of the semantic embeddings of stage 3 to the visual space via the utilization of the KG of step 2, the visual embeddings of stage 3 and the trained GNN of stage 4.  
  \item Incorporation of the projected embeddings of stage 5 into a pre-trained Visual Classifier.  
\end{enumerate}

\subsubsection{Stage 1: Prompting \& semantic representation.}

\vspace*{0.1cm}\noindent\textbf{Prompting:}
The prompts\footnote{\label{note1}The prompts are available at  \url{https://docs.google.com/spreadsheets/d/11jo4wSdBrI-rj3jI-UpU39KDfBGGDhsxIEs9UUShh1w/edit?usp=sharing}} that were directed to the LLM can be roughly categorized into two groups. The first group contains prompts that examine the various objects as items of the physical world and emphasize on the description of their properties. The second group focuses on the visual characteristics of the objects and how they are depicted in images or described in image captions. 

More specifically, for the first group some of the prompts are related only to the classes of ImagetNet (1K version)  e.g., {\sl `For the concept [ImageNet class] give me a short narration of about 160 words'}. Other prompts combine the classes of the ImageNet with a set of predefined states, e.g., {\sl `make a connection between the concept [ImageNet class] and [state1, ..., stateN]'}, while others try to find asymmetries e.g., {\sl `What action can I do with an object in state [state] in the physical world that I cannot do when the object is in [opposite state]?'} and causal chains between pair of states e.g., {\sl `Give me 10 examples of how a human can take an [state] object of the physical world and make it [opposite state]'}.

Most of the prompts follow a zero-shot prompting style but in-context learning in the form of few-shot prompting e.g., {\sl `What objects of the physical world other than charger, phone, socket can be also connected?'} and chain-of-thought via the simplistic addition of {\sl `Let's consider it step by step.'} at the end of the prompts, have also applied. The second group consists of prompts like {\sl `Describe what a(n) [ImageNet class] looks like?'}, {\sl 
`Describe a photo of a(n) [ImageNet class]'}, and {\sl `Give me a caption of an image of a(n) [ImageNet class]'}.  

\vspace*{0.1cm}\noindent\textbf{Corpus processing:}
Initially, the responses of the model were cleaned up from the text that started the conversation. In most of the cases, the text follows a  standardized format e.g.,  {\sl `As a helpful and respectful assistant, I would be happy to assist you'}, {\sl `Sure, I'd be happy to help!'} etc. Table~\ref{tab:corporaoverview} shows an overview of the corpora. The first corpus (Basic) corresponds to the first group of prompts while the second (Extended) to the unification of the two groups of prompts. 

\begin{table}
\centering
\begin{tabular}{l|r|r|r}
\hline
{\bf Corpus }&{\bf Total }& {\bf Unique}  & {\bf Vocab} \\ \hline
Basic& 665.594     &  13.688 & 7.691  \\ 
Extended& 1.573.843     &  19.183 & 11.372  \\   \hline
\end{tabular}
\caption{Overview of the two corpora utilized in our work w.r.t. number of total, unique and vocab tokens. The vocabulary for both corpora consists of unique tokens with a minimum frequency equal to 3.}
\label{tab:corporaoverview}
\end{table}

\vspace*{0.1cm}\noindent\textbf{LLM-based and generic embeddings:}
For the training of the LLM-based static word embeddings (domain) the  
Global Vectors (GloVe)~\cite{Pennington2014glove}, the fastText~\cite{bojanowski2017enriching} and the word2vec~\cite{Mikolov2013word2vec}  algorithms were used. 
As general-purpose embeddings, we employed pre-trained word embeddings of the GloVe\footnote{\url{https://github.com/stanfordnlp/GloVe} (6B and 42B versions)}, the fastText\footnote{\url{https://fasttext.cc/docs/en/english-vectors.html}} and the word2vec\footnote{\url{https://github.com/harmanpreet93/load-word2vec-google}} models. In addition, the ConceptNet Numberbatch\footnote{\url{https://github.com/commonsense/conceptnet-numberbatch}} KG-enhanced word embeddings were also used which retrofit~\cite{Faruqui2014retrofit} word2vec and GloVe word representations to the ConceptNet graph~\cite{speer2017conceptnet}.

Each type of embeddings was employed both separately and in conjunction with other embeddings types. For the combination of the embeddings of different type we experimented  with a number of different fusion techniques. The  goal was  to exploit both the robustness of the generic embeddings and  the domain-specific semantic information that is captured by the LLM-based embeddings. 
Below, we briefly describe the undertaken fusion approaches:

\begin{itemize} 
\item \textbf{Averaging:} The arithmetic mean of the two-word embedding sets is calculated following  \citet{coates2018averaging}.
\item \textbf{Concatenation:} The LLM-based embeddings were concatenated with the generic embeddings for the common words between the two vocabularies. The intuition for this setting is that the model during training will prioritize certain embedding dimensions.
\item \textbf{PCA:} The concatenated embeddings were linearly transformed via the PCA technique to preserve the relevant information by approximating the input vectors and minimizing the number of components. A z-means normalization was applied beforehand and we kept the top-300 components.
\item \textbf{UMAP:} In this case, the UMAP~\cite{mcinnes2020umap}   which is a non-linear dimension reduction technique, was employed over the concatenated embeddings. UMAP has previously been applied for dimensionality reduction on text embeddings~\citep{Massarenti2021DialogCA}.
\item \textbf{Autoencoder:}  An auto-encoder~\citep{Vincent2008autoencoders} was employed taking as input the concatenated embeddings. For each word, we consider the vector of values produced by the latent layer as the combined word embedding.
\end{itemize}

\subsubsection{Stage 2: KG construction.}

In this stage, a Knowledge Graph is built, with its topology tailored to the target classes. The process involves querying one or more commonsense knowledge sources to retrieve neighbors for each target class, up to a specified hop threshold. The KG is then constructed by merging all query results. Intentionally, no filtering or other post-processing has been performed on the query results, to ensure a rapid and automatic execution of the entire stage. As such, it is important to note that this approach unavoidably introduces noise into the constructed KG.

\subsubsection{Stage 3: Production of Semantic and Visual Embeddings.}
The corpus generated by the LLM in the initial stage undergoes processing, leading to the generation of semantic representations in the form of word embeddings. These domain-specific embeddings become pivotal in subsequent stages, either utilized in isolation or in conjunction with general-purpose pre-trained embeddings. To generate visual embeddings, a Convolutional Neural Network (CNN) classifier is trained on images, and the last layer's learned weights serve as visual embeddings corresponding to the classes of the images used for training.

\subsubsection{Stage 4: GNN Training.}
A Graph Neural Network (GNN) is trained to effectively project semantic embeddings onto the visual space. Specifically, the GNN takes as input a KG and semantic embeddings corresponding to each node in the KG. The training procedure involves learning to generate visual embeddings for each node. The visual embeddings used to guide the GNN training are those obtained in stage 3.
The significance of the GNN lies in its ability to capture graph dependencies through message passing between nodes, as elucidated by~\cite{Zhou2020}. With this approach, the GNN is designed to learn the interrelations between nodes in  KGs. In our scenario, this involves the KG that is provided as input to the GNN during the training process and the KGs that are provided during the embedding projection phase that is described next. 


\subsubsection{Stage 5: Embeddings Projection to the Visual Space.}
The GNN, trained in stage 4, is fed with the KG constructed in stage 2 and the semantic embeddings generated during stage 3. The GNN processes this input and produces visual embeddings corresponding to the target classes. This step ensures that the GNN effectively leverages both the structure of the KG and the semantic information encoded in the embeddings to generate meaningful visual representations for the specified target classes.

\subsubsection{Stage 6: Zero-shot classifier adaptation.}
In the final step, the last layer of the  CNN classifier, initially utilized as a guide in stage 3, is replaced with the visual embeddings generated in the previous stage. This adaptation enables the modified CNN to effectively discern and identify the target classes, leveraging the refined visual embeddings for improved classification performance.

\section{Experimental Evaluation}
We conducted a series of experiments with a dual objective. Our primary goal was to explore crucial aspects related to the integration of LLMs into the Object Agnostic Zero Shot Classification task pipeline. To achieve this, we carried out a comprehensive ablation study covering various parameters. Additionally, we aimed to evaluate the performance of the best-ablated model against existing competing methods. In what follows, we provide detailed insights into the implementation and results for both experimental settings.


\begin{table}[t]
\centering
\begin{tabular}{l|c|c|c|c|c}
\hline
\textbf{Combination} & \textbf{CN}  & \textbf{VG} & \textbf{WN}  & \textbf{WK} & \textbf{AT}\\ \hline
KG1& \checkmark    &   &  & & \\ 
KG2&    & \checkmark   &  & & \\ 
KG3& \checkmark &\checkmark    &   &   \\ 
KG4& \checkmark    &   &\checkmark &  &  \\ 
KG5& \checkmark    &    &  &\checkmark &  \\ 
KG6& \checkmark    &    &  \checkmark  &\checkmark &  \\ 
KG7& &\checkmark      &\checkmark &  &  \\ 
KG8& &\checkmark        &  &\checkmark &  \\ 
KG9& &\checkmark        &  \checkmark  &\checkmark &  \\ 
KG10&  \checkmark &\checkmark        &  \checkmark  &\checkmark &  \checkmark \\  \hline

\end{tabular}
\caption{Details of the KGs used for the production of the visual embeddings (stage 5 of methodology).  Each  KG combination is constructed for three different hops thresholds:  1,2 and 3. {CN: ConceptNet}, {VG: Visual Genome}. {WN: WordNet}, {WK: WikiData}, {AT: Atomic}.}
\label{tab:KGs}
\end{table}

\subsection{Implementation \& Evaluation Issues}

The LLM utilized in our study is Llama2~\citep{Touvron2023llama}, an updated version of LLama1. Llama2 belongs to the family of decoder-only large language models, comprising a suite of pre-trained and fine-tuned generative text models ranging in scale from 7 billion to 70 billion parameters. The family of Llama2 models is freely accessible for both research and commercial purposes\footnote{\url{https://about.fb.com/news/2023/07/llama-2}}. For the purposes of this research, we specifically employed the 13 billion parameters variant of the Llama-2-Chat model. In particular, a quantized model provided by TheBloke\footnote{The llama-2-13b.Q5-K-M.gguf file at \url{https://huggingface.co/TheBloke/Llama-2-13B-GGUF}} was utilized. The selection of the Llama 2-Chat model was made based on its generally superior performance compared to other available open-source chat models. The predefined values for temperature and top\_p parameters were maintained at 0.6 and 0.9, respectively.

\begin{table*}
\centering
\begin{tabular}{l|c|c|c|c}
\hline
{\bf Corpus}& {\bf OSDD}   & {\bf CGQA-States}       & {\bf MIT-States}       & {\bf VAW}      \\ \hline
Bas Domain-No & 19.60  $\pm$  2.00  &31.07  $\pm$  3.51   &29.54  $\pm$   2.90   &17.93  $\pm$   2.88    \\ 
Bas Domain& \bf 24.64  $\pm$  2.01  &36.32  $\pm$  2.23   & \bf 37.80  $\pm$   3.68   & \bf 24.52  $\pm$   3.13    \\ 
Ext Domain& 18.83  $\pm$  1.52  &  \bf 38.59  $\pm$  4.33   &36.53  $\pm$   5.00   &19.79  $\pm$   1.26    \\   
          \hline
\end{tabular}
\caption{Results for different corpora. Bas Domain-No: Basic corpus without stopwords. Bas Domain: Basic corpus with stopwords. Ext Domain: Extended corpus with stopwords. In all cases, the GloVe representation was employed as the chosen method for encoding the textual information within the respective corpus.}
\label{tab:ab1}
\end{table*}

\begin{table*}
\centering
\begin{tabular}{l|c|c|c|c}
\hline
{\bf Semantic Source}& {\bf OSDD}   & {\bf CGQA-States}       & {\bf MIT-States}       & {\bf VAW}  \\ \hline
GloVe 6B &23.55  $\pm$  0.61  &37.54  $\pm$  3.07   &\bf  46.58  $\pm$   4.17   & \bf 25.45  $\pm$   2.54    \\ 
GloVe 42B & 21.97  $\pm$  2.78  &37.81  $\pm$  4.61   &41.78  $\pm$   3.96   &17.97  $\pm$   3.21    \\ 
fastText 16B  & 23.42  $\pm$  1.90  & \bf  40.12  $\pm$  4.31   & 42.67  $\pm$   2.75   &23.64  $\pm$   1.97    \\ 
Word2Vec& 23.42  $\pm$  1.90  &40.12  $\pm$  4.31   &42.67  $\pm$   2.75   &23.64  $\pm$   1.97 \\ 
Numberbatch & \bf  24.29  $\pm$  1.35  &26.56  $\pm$  2.19   &36.48  $\pm$   2.60   &19.08  $\pm$   2.39    \\ 
\hline
GloVe Bas Domain&   \bf 24.64  $\pm$  2.01  &36.32  $\pm$  2.23   &37.80  $\pm$   3.68   & \bf  24.52  $\pm$   3.13    \\ 
fastText Bas Domain& 23.41  $\pm$  2.61  &  \bf 41.54  $\pm$  2.57   &\bf  40.92  $\pm$   2.41   &24.23  $\pm$   2.84    \\ 
Word2Vec Bas Domain& 24.11  $\pm$  2.25  &42.01  $\pm$  2.51   &42.05  $\pm$   2.61   &24.21  $\pm$   2.75    \\     
          \hline
\end{tabular}
\caption{Results for different types of semantic embeddings. GloVe 6B: 6B tokens trained on Wikipedia 2014 and Gigaword 5. GloVe 42B: 42B tokens trained on Common Crawl.  fastText 16B: 16B tokens trained on Wikipedia 2017, UMBC webBase corpus, and statmt.org news dataset. Word2Vec: 100B tokens trained on a part of the Google News dataset. Numberbatch: retrofitted word2vec-100B  trained on a part of the Google News dataset
and GloVe-840B trained on Common Crawl word representations to the ConceptNet 5.5 graph.}
\label{tab:ab2}
\end{table*}

\begin{table*}
\centering
\begin{tabular}{l|c|c|c|c}
\hline
{\bf Semantic Source}& {\bf OSDD}   & {\bf CGQA-States}       & {\bf MIT-States}       & {\bf VAW}  \\ \hline
GloVe 6B &23.55  $\pm$  0.61  &37.54  $\pm$  3.07   &\bf  46.58  $\pm$   4.17   &25.45  $\pm$   2.54    \\
GloVe 6B + Bas Domain& 24.77  $\pm$  2.27  &42.24  $\pm$  2.93   &42.58  $\pm$   2.35   &25.06  $\pm$   2.61    \\ 
GloVe 6B + Ext Domain& \bf  25.23  $\pm$  2.54  &\bf  43.11  $\pm$  4.23   &42.43  $\pm$   4.62   &\bf  25.97  $\pm$   2.19    \\ 
GloVe 6B +  Ran& 21.51  $\pm$  1.22  &29.78  $\pm$  4.59   &43.93  $\pm$   3.00   &20.95  $\pm$   1.90    \\   \hline
GloVe 6B - BDCVR & 23.17  $\pm$  1.73  &37.30  $\pm$  3.97   &43.58  $\pm$   2.23   &22.64  $\pm$   1.44    \\
         
\hline
GloVe 42B  & 21.97  $\pm$  2.78  &37.81  $\pm$  4.61   &41.78  $\pm$   3.96   &17.97  $\pm$   3.21    \\ 
GloVe 42B + Bas Domain& 23.11  $\pm$  2.93  &39.31  $\pm$  4.54   &42.96 $\pm$   4.14   &20.14  $\pm$   3.74    \\ 
GloVe 42B + Ext Domain& \bf 23.64  $\pm$  2.78  & \bf  40.47  $\pm$  4.59   & \bf  43.92 $\pm$   4.35   & \bf  22.31  $\pm$   3.54    \\ 
GloVe 42B + Ran & 19.54  $\pm$  2.14  &30.14  $\pm$  2.84   &38.54 $\pm$   4.02   &15.56  $\pm$   2.74    \\ 

          \hline
\end{tabular}
\caption{Results for different concatenation combinations of domain-specific and general-purpose embeddings. In all cases, the concatenated embeddings have a dimensionality of 600. 
Ran:  Random Glove embeddings. BDCVR: The general-purpose embeddings have the same coverage as Bas Domain embeddings. For explanations regarding the remaining abbreviations, please refer to the previous tables.
}
\label{tab:ab3}
\end{table*}

\begin{table*}
\centering
\begin{tabular}{cc|c|c|c|c}
\hline
  \multicolumn{2}{c|}{ \bf Method} & {\bf OSDD}   & {\bf CGQA-States}       & {\bf MIT-States}       & {\bf VAW}   \\ \hline
 \multirow{5}{*}{GloVe 6B + Bas Domain}&AVG & 23.91  $\pm$  1.65  &37.88  $\pm$  3.23   &41.42  $\pm$   3.50   &23.19  $\pm$  1.73 \\ 
&Concat& 24.77  $\pm$  2.27  & \bf 42.24  $\pm$  2.93   &42.58  $\pm$   2.35   &25.06  $\pm$   2.61    \\
&PCA & \bf 27.27  $\pm$  1.23  &37.45  $\pm$  3.54   &\bf 45.22  $\pm$   6.50   &25.15  $\pm$   3.15    \\ 
&UMAP& 25.14  $\pm$  1.58  &34.86  $\pm$  3.06   &30.06  $\pm$   2.84   &\bf 25.32  $\pm$   2.80    \\
&Autoencoder & 24.69  $\pm$  1.82  &38.77  $\pm$  4.20   &42.64  $\pm$   5.10   &24.57  $\pm$   2.55    \\  \hline

 \multirow{5}{*}{GloVe 6B + Ext Domain}&AVG & 24.33  $\pm$  1.82  &39.08  $\pm$  3.42   &42.87 $\pm$   3.47   &23.54 $\pm$  1.92 \\ 
&Concat& 25.68  $\pm$  2.65  &40.12  $\pm$  2.47   &41.35  $\pm$   2.11   &27.14  $\pm$   2.61    \\
&PCA & 26.56  $\pm$  2.90  & \bf 42.94  $\pm$  4.07   &42.08  $\pm$   4.57   &25.33  $\pm$   2.69    \\ 
&UMAP& 27.41  $\pm$  1.78  &36.41  $\pm$  3.42   &32.81 $\pm$   2.79   &26.49  $\pm$   2.30    \\
&Autoencoder & \bf 28.55  $\pm$  1.46  &41.53  $\pm$  4.45   &\bf 49.02  $\pm$   3.13   &\bf 29.29  $\pm$   3.64    \\ 
\hline

\end{tabular}
\caption{Results for different fusion methods of domain-specific and general-purpose embeddings. AVG: The two embeddings were averaged. Concat: The two types of embeddings were concatenated resulting in vectors of dimension  600. PCA:  Principal Component Analysis is performed on the concatenated embeddings, producing a vector of dimension 300. UMAP: Uniform Manifold Approximation and Projection is performed to the concatenation of the two embeddings, producing a vector of dimension 300. Autoencoder: An autoencoder is employed, with the concatenated embeddings serving as input and a vector of dimension 300 is produced.  For explanations regarding the remaining abbreviations, please refer to the previous tables.}
\label{tab:ab4}
\end{table*}

\begin{table*}
\centering
\begin{tabular}{cc|c|c|c|c}
\hline
 \bf{Sources/Technique}&{\bf CNN Model} & {\bf OSDD}   & {\bf CGQA-States}       & {\bf MIT-States}       & {\bf VAW} \\ \hline

 \multirow{2}{*}{Bas Domain }&RN101    & 24.64  $\pm$  2.01  &\bf 36.32  $\pm$  2.23   & \bf 37.80  $\pm$   3.68   &24.52  $\pm$   3.13      \\ 
& ViT-16  & \bf 25.26  $\pm$  1.94  &34.91  $\pm$  6.40   &32.03  $\pm$   2.88   & \bf 26.75  $\pm$   3.99    \\ 
  \hline
 
 \multirow{2}{*}{GloVe 6B + Bas Domain (PCA) } & RN101 &  \bf  27.27  $\pm$  1.23  & \bf 37.45  $\pm$  3.54   &\bf 45.22  $\pm$   6.50   & \bf 25.15  $\pm$   3.15    \\ 
&ViT-16 (PCA)& 25.63  $\pm$  1.59  &30.08  $\pm$  2.94   &39.24  $\pm$   4.30   &23.09  $\pm$   1.41    \\  \hline

 \multirow{2}{*}{GloVe 6B + Bas Domain (UMAP) }  & RN101 &    25.14  $\pm$  1.58  &\bf 34.86  $\pm$  3.06   & \bf 30.06  $\pm$   2.84   &25.32  $\pm$   2.80    \\ 
& ViT-16 & \bf 27.09  $\pm$  2.09  &30.65  $\pm$  2.39   &24.84  $\pm$   2.28   &\bf 25.75  $\pm$   2.53    \\  \hline  
 \multirow{2}{*}{GloVe 6B + Ext Domain (PCA) } & RN101     & 26.56  $\pm$  2.90  &\bf 42.94  $\pm$  4.07   &\bf 42.08  $\pm$   4.57   &\bf 25.33  $\pm$   2.69    \\ 
& ViT-16 (PCA-Aug)& \bf27.14  $\pm$  1.26  &35.01  $\pm$  2.30   &35.10  $\pm$   3.70   &\bf 25.33  $\pm$   2.57    \\     \hline

\end{tabular}
\caption{Results for the two CNN Models that are used in the final stage of classification and from which the visual embeddings are extracted in stage 2. RN101: ResNet-101. ViT-16: Vit-16/B. For an explanation regarding the remaining abbreviations please refer to previous tables.}
\label{tab:ab5}
\end{table*}

\begin{table*}[ht]
\centering
\begin{tabular}{|l|c|c|c|c|}
\hline
\multirow{1}{*}{{\bf Method}} & {\bf OSDD}                                                          & {\bf CGQA-States}                                                   & {\bf MIT-States}                                                    & {\bf VAW}                                                           \\ \cline{1-5}

  Ours & \textbf{33.0} & \textbf{48.1} & \textbf{53.7} & \textbf{32.7} \\
                                         \hline

OaSc  &29.1 & 39.4& 47.4 & 25.4 \\ 
                                         \hline\hline

CLIP-RN101                               & 22.5                                                          & 46.9                                                 &        39.3                                                         & 28.0                                                         \\ \hline

CLIP-VITBP16                                & 28.8                                        &                 {44.9}                                                          & {46.4}                                                         & {30.1} \\  \hline

CLIP-VITLP14                                & 28.4                                       &                43.4                                                          & 48.6                                                         & 27.9   \\ \hline
                                                      
ALIGN                                    & {29.5}                                                       & 40.0                                                          & 44.2                                                          & 28.4                                                          \\ \hline
BLIP                                     & 13.3                                                          & 26.0                                                          & 27.2                                                          & 16.1                                                          \\ \hline \hline
 RN101 (Supervised) & 67.5             & 60.5                                &85.3                       & 51.9              \\ \hline
\end{tabular}
\caption{Experimental results of the proposed approach for the zero-shot object state classification task. The performance reported refers to the following four datasets:  OSDD~\cite{gouidis2022}, CGQA-States~\cite{naeem2021learning}, MIT-States~\cite{Isola2015}, VAW~\cite{Pham2021CVPR}. As a frame of reference, the performance of a supervised visual state classification model is included. This reference model relies on the ResNet-101 network architecture, trained in a fully supervised setting on each dataset independently. }
    \label{tab:exps}
\end{table*}

The KG utilized for the projection from semantic space to visual space is derived from the ImageNet Graph, organized according to the WordNet hierarchy. This KG corresponds to the ImageNet1K dataset, comprising approximately 1,350,000 images across 1000 different classes~\cite{krizhevsky2012imagenet}. The GNN model is trained from scratch, following the methodology outlined in~\cite{Kampffmeyer2019}, for 1500 epochs. The architecture of the network is convolutional-based and is inspired by \cite{nayak2022learning}.
During training, 950 classes were randomly selected from the ImageNet (ILSVRC 2012) dataset \cite{russakovsky2015imagenet}, while the remaining 50 classes were reserved for validation purposes. Two widely used pre-trained classifiers, namely ResNet-101~\cite{Gong2022ResNet10} and ViT-B/16~\cite{dosovitskiy2021image}, are employed in our experiments.

For the generation of visual embeddings, five commonsense sources were queried to construct KGs: ConceptNet~\cite{speer2017conceptnet}, WordNet~\cite{Miller1994Wordnet}, Visual Genome~\cite{Krishna2017visualgenome}, Wikidata~\cite{vrandevcic2014wikidata}, and Atomic~\cite{Sap2019Atomic}. In total, 30 different KGs were constructed, representing 10 combinations of sources across three hop distances: 1, 2, and 3. Table~\ref{tab:KGs} provides detailed information about these KGs.

Concerning semantic embeddings, we experimented with four different approaches for semantic representation: Glove, Word2Vec, fastText, and Numberbatch. For both types of embeddings (domain and general purpose) a dimension ($d$) of 300 is used, which is a common choice for word embeddings. In addition, previous work~\citep{Lai2016plateau} has shown that the performance of embeddings tends to plateau after a certain dimension i.e., 100. In respect to the preparation of corpus for the training of word embeddings, the segmentation of the sentences, and the  
tokenization was performed with the StanfordCoreNLP~\citep{Qi2018stanfordcorenlp} and all tokens were lowercased.

\vspace*{0.1cm}\noindent\textbf{Datasets}: 
Except for the OSDD dataset~\cite{gouidis2022}, which is explicitly designed for the state detection task, currently, no other dataset exists exclusively dedicated to capturing object states in images. Nonetheless, certain pre-existing datasets focused on object detection and classification, incorporate object states as a subset within their array of object classes. One such dataset is the Visual Attributes in the Wild (VAW) dataset~\cite{Pham2021CVPR}, encompassing object state classes as a distinct subset within its attribute annotations. Similarly, MIT-States~\cite{Isola2015} and CGQA-States~\cite{Mancini2022} are widely employed datasets within the domain of attribute classification.



\vspace*{0.1cm}\noindent\textbf{Metrics}: 
Our evaluation protocol adheres to the established zero-shot evaluation method as outlined in~\cite{purushwalkam2019task}. In contrast to the conventional setting where accuracy across all classes is typically reported, our approach involves computing accuracy for each class individually. Subsequently, an overall mean average is calculated across these individual class results. This method assigns equal weight to each class, irrespective of the respective number of samples associated with each class.


\subsection{Ablation Study}
The ablation study focuses on addressing the following key questions:

\begin{itemize}
    \item Can domain-specific knowledge generated by an LLM contribute to resolving the zero-shot problem under examination?
      \item  What is the optimal representation method for domain-specific knowledge generated by an LLM?
    \item  Can domain-specific knowledge from the LLM be effectively combined with general-purpose knowledge derived from large corpora, and what is the optimal approach for this integration? 
        \item To what extent does the size of the data used for obtaining general-purpose knowledge impact the performance of zero-shot methods?
 \end{itemize} 

 Tables~\ref{tab:ab1}-~\ref{tab:ab5} present the results of the ablation study with the reported scores in each table corresponding to the mean values and standard deviations of the performance across the 30 different KGs. In more detail, Table~\ref{tab:ab1} shows results related to the corpus size used as input for the LLM. Table~\ref{tab:ab2} contains the results for the impact of the representation used for the semantic vectors. Table~\ref{tab:ab3} presents results concerning different combinations of domain-specific and general-purpose knowledge. Lastly, Tables~\ref{tab:ab4} and ~\ref{tab:ab5} pertain to experiments involving the dimensionality reduction of the semantic vectors and the CNN model used as a classifier, respectively. 




%


\vspace*{0.1cm}\noindent\textbf{Impact of corpus size}:
Analyzing the impact of corpus size (Table~\ref{tab:ab1}), it is observed that the best results are achieved with the basic-sized corpus rather than the extended one. This can be attributed to the domain specificity of prompts used for the creation of the basic corpus since the prompts used for extending the corpus are more general. Additionally, in line with expectations from the literature~\cite{lison-kutuzov-2017-redefining, Rahimi2022stopwords} that shows that stopword removal is not beneficial in all downstream tasks, excluding stopwords from the basic corpus results in a decrease in performance.
 

\vspace*{0.1cm}\noindent\textbf{Types of semantic embeddings}:
Analyzing the results related to different types of semantic embeddings (Table~\ref{tab:ab2}), the following observations can be made. For general-purpose embeddings, the best scores are achieved by GloVe 6B and fastText 16B, which have been trained on fewer tokens. This outcome supports the notion that embeddings from large corpora may be tainted with noise, thereby reducing performance. Regarding domain-specific embeddings generated by the LLM corpus, the best representation is GloVe, followed by fastText. Moreover, it is observed that domain-specific embeddings achieve performance comparable to that of general-purpose embeddings across all representation classes.


\vspace*{0.1cm}\noindent\textbf{Fusion of domain-specific / general-purpose embeddings}:
The key findings from the fusion of domain-specific and general-purpose embeddings presented in Table~\ref{tab:ab3} are as follows. The concatenation of the two types of embeddings leads to performance gains in most cases for GloVe 6B (except for the CGQA-States dataset) and in all cases for GloVe 42B. These gains can be attributed to domain-specific knowledge: fusion with random embeddings results in a significant decrease in performance. Moreover, the version of GloVe 6B embeddings that has the same coverage as the embedding of the basic corpus (GloVe 6B - BDCVR in the Table) scores lower than GloVe 6B. This result indicates that the combination of domain-specific and general-purpose embeddings achieves better performance \textit{despite} their lower coverage.


\vspace*{0.1cm}\noindent\textbf{Impact of dimensionality reduction}:
Analyzing the results for dimensionality reduction and fusion techniques in Table~\ref{tab:ab4}, it is observed that averaging is the least successful technique, achieving the worst results in the majority of cases. In the fusion of GloVe 6B with the basic corpus, the best scores are obtained with PCA, whereas Autoencoder achieves the best result in the combination of GloVe 6B with the extended corpus. Finally, results regarding the CNN model used as the State Classifier (Table~\ref{tab:ab5}) suggest that ResNet-101 outperforms ViT-16/B in most cases.

\subsection{Experimental Results}
To the best of our knowledge, aside OaSc \cite{gouidis204}, there currently is no readily available object-agnostic state model suitable for deployment in a zero-shot setting. Therefore, in addition to OaSc,   we have chosen to employ additionally three state-of-the-art Vision Language Models (VLMs) that support this functionality: CLIP~\cite{radford2021learning}, ALIGN~\cite{jia2021scaling}, and BLIP~\cite{li2022blip}. In our experimentation, we utilize three variants of CLIP one of ALIGN and one of BLIP. It is essential to note that all these models indirectly violate the fundamental assumptions of the zero-shot setting, as the training data includes pairs of text and images containing the target classes relevant to our task.


Table~\ref{tab:exps} presents the results of the experimental evaluation across the four datasets.  The reported scores for our method correspond to the performance of the best ablated model, which utilized the fusion of LLM-based embeddings from the extended corpus with GloVe 6B pre-trained embeddings. This model employed the Autoencoder as a dimensionality reduction technique and ResNet-101 as the classifier. Regarding the VLMs, following the typical procedure~\cite{xue2022clip}, we used 16 different prompts\textsuperscript{\ref{note1}} and report the mean value of these scores.

The results indicate that our model not only outperforms OaSc by a significant margin but also achieves superior performance compared to every competing VLM across all datasets. These findings strongly support the robustness of our approach, especially considering that the competing VLMs are recognized state-of-the-art models for various zero-shot tasks.

\section{Conclusions and Future Work}

This work represents a step towards integrating an LLM into a hybrid framework for addressing a crucial Computer Vision problem. The experimental results strongly suggest that incorporating knowledge from the LLM can substantially enhance the performance of the framework. The success of this integration instills confidence in further exploration of ideas for leveraging LLMs in the near future.

Firstly, we consider that the utilization of the LLMs can be extended to the construction and refinement of KGs used for generating the visual embeddings essential for the zero-shot classification task. Specifically, by employing tailored prompts, the LLM can generate text, which, after parsing, will yield a KG. Similarly, querying the LLM about an existing KG could facilitate the addition or pruning of edges. Furthermore, the LLM can play a role in constructing the KG used for training the GNN which is essential for the projection of the semantic embeddings to the visual space.

In this work we utilize a 13B parameter model to strike a reasonable balance between cost and performance, The obtained results are highly promising as they already advance the state-of-the-art performance on the problem. However, employing a more robust model might enhance the capture of commonsense knowledge and human reasoning capabilities. Additionally, searching for the optimal set of prompts remains an open subject, as it plays a crucial role in determining the quality of the LLM-based embeddings. Finally, we aim to further enhance the capabilities of the LLM model by fine-tuning it using publicly available image-text datasets, such as Visual Genome.

\section{Acknowledgements}
The Hellenic Foundation for Research and Innovation (H.F.R.I.) funded this research project under the 3rd Call for  H.F.R.I. Research Projects to support Post-Doctoral Researchers (Project Number 7678 InterLinK: Visual Recognition and Anticipation of Human-Object Interactions using Deep Learning, Knowledge Graphs and Reasoning) and under the “1st Call for H.F.R.I Research Projects to support Faculty members and Researchers and the procurement of high-cost research equipment”, project I.C.Humans, number 91.





\bibliography{aaai24}

\end{document}